\documentclass[12pt,sort,compress]{article}

\usepackage{arxiv}

\usepackage{comment}
\usepackage{soul} % to strikethrough
\usepackage{amssymb,amsmath}
\usepackage{mathtools}
\usepackage{algorithm,algpseudocode}
\usepackage{xifthen}
\usepackage[table]{xcolor}
\usepackage{booktabs}
\usepackage{makecell}
\usepackage{tabularx}
\usepackage{graphics,graphicx}
\usepackage[inkscapelatex=false]{svg}
\usepackage{tikz}
\usetikzlibrary{shapes.geometric, shapes.arrows, angles, quotes, arrows, patterns, decorations, positioning}
\usepackage{pgfplots}
\pgfplotsset{compat=1.6}
\usepackage{pifont}
\usepackage{lipsum}
\usepackage{mwe}
\usepackage{graphicx}
\usepackage[percent]{overpic}
\newcolumntype{P}[1]{>{\centering\arraybackslash}p{#1}}

\graphicspath{{./FIGS/}}

\usepackage[labelformat=simple]{subcaption}
\usepackage{lineno}
\usepackage{natbib}
\usepackage{hyperref}
\hypersetup{
    colorlinks=true,
    pdfauthor=author %Prevents 3 very obscure hyperref warnings
}
\usepackage{siunitx}

\newcommand{\fig}[1]{Figure~\ref{#1}}

\newcommand{\tab}[1]{Table~\ref{#1}}
\newcommand{\sect}[1]{Section~\ref{#1}}

\newcommand{\pmc}{$\text{PM}_{10}$}
\newcommand{\pmf}{$\text{PM}_{2.5}$}
\newcommand{\nod}{$\text{NO}_{2}$}
\newcommand{\sod}{$\text{SO}_{2}$}
\newcommand{\oz}{$\text{O}_{3}$}
\newcommand{\nox}{$\text{NO}_{x}$}

\newcommand{\ie}{i.e.,}

\newcommand{\colr}[1]{{\color{black}#1}}

\title{Air Quality Station Simulation via LSTM and Attention-Based Modelling}

\author{
 Alexander Kostadinov \\
  GATE Institute\\
  Sofia University ``St Kliment Ohridski''\\
  5 James Bourchier Blvd, 1164, Sofia\\
  \texttt{alexander.kostadinov@gate-ai.eu} \\
 \And
 Petar O. Hristov \\
  GATE Institute\\
  Sofia University ``St Kliment Ohridski''\\
  5 James Bourchier Blvd, 1164, Sofia\\
  \texttt{petar.hristov@gate-ai.eu} \\
  \And
 Dessislava Petrova-Antonova \\
  GATE Institute\\
  Sofia University ``St Kliment Ohridski''\\
  5 James Bourchier Blvd, 1164, Sofia\\
  \texttt{dessislava.petrova@gate-ai.eu} \\
}

\begin{document}
    \maketitle
    \begin{abstract}
    Poor air quality in urban areas is driven by a complex chain of processes and presents a significant public health concern.
    To better understand and control the mechanisms that determine air quality, cities deploy networks of measurement stations, and launch initiatives for collecting denser data about the concentration
    of pollutants in the atmosphere. Extracting insights from the stations relies on their reliable and uninterrupted operation. However, hardware is susceptible to faults and black-outs that may result in data unavailability, which affects the overall quality of analyses.
    
    In this paper, we present a deep-learning model, %spatio-temporal attention, dual long short-term memory model, 
    called SATADL, which can infer complex relations and output multiple-hour-ahead air-quality forecasts. The goal of the model is to simulate the  measurements of an  unresponsive station until its operation is restored. The architecture of the model, which allows it to extract information from different aspects of the data, is described in detail and a careful examination of all of its components is provided. 
    
    We demonstrate the performance of SATADL on four sets of air quality stations from around the world, by using it to simulate the concentration of \pmc\ for periods of hypothetical failures of one of the measurement stations, lasting for as long as 48 hours. A selection of baseline and published deep learning models were trained and used as a benchmark. The results show that SATADL performs better across different prediction windows, for both coefficient of determination and root mean squared error, demonstrating its suitability as a virtual proxy station.
    \end{abstract}
    

\section{Introduction}
    \subsection{Background and motivation}
    The quality of urban air is driven by a sophisticated chain of processes of both anthropogenic and natural origin, among others. Air pollution has become a major problem worldwide and endangers the people's health and the environment. Poor air quality leads to multiple diseases and complications \cite{Kampa:2008}, with an estimation of \num{6.7} million deaths globally, according to the World Health Organisation (WHO)%
    \footnote{According to the official website of WHO, accessible at https://www.who.int/news-room/fact-sheets/detail/ambient-(outdoor)-air-quality-and-health}.
    Because of the complexity and severity of the problem, the WHO%
    \footnote{The interested reader is redirected to the ``\textit{Estimating the Morbidity from Air Pollution and its Economic Costs (EMAPEC)}'' project by WHO.}
    is continually seeking ways to improve how risk assessments are conducted \cite{Forastiere:2024}.
    
    Because air pollution is related to many factors of the urban design and development and because it has a significant impact on the liveability of cities, the quality of air is used as a natural progress indicator towards the transformation of urban areas into smart cities. In smart cities traditional infrastructure and public services are optimised through digital solutions, which in turn, enhances the quality of life for its residents and boosts economic efficiency. A driving concept for smart cities are the so-called urban digital twins \cite{Abdelrahman:2025}, which serve as digital replicas of different dimensions and processes of the city. One can argue that, for the reasons laid out above, an urban digital twin is in fact a digital twin for air quality!

    To address the dangers of poor air quality, municipalities around the world have established monitoring systems that measure the concentrations of pollutants, such as particulate matter no larger than \qty{10}{\micro\meter} in aerodynamic diameter (\pmc), nitrogen dioxide (\nod), sulphur dioxide (\sod) and others, as well as meteorological factors like atmospheric temperature and pressure, wind speed and direction, and rainfall intensity, which contribute to the dispersion of pollutants. Despite providing a reference for the concentration levels of different air pollutants , these systems, as any other piece of hardware are susceptible to technical issues, weather events, and power outages, that compromise their reliability as measurement devices. In addition, certified air quality stations are expensive equipment and are usually distributed at only a few carefully chosen locations, making each individual station critical to understanding the quality of air in the city.   

    To translate monitoring into actionable insights, authorities need ways to analyse spatial, as well as past and future trends of pollutants, tasks which require the continuous availability and aggregation of data from all devices deployed in the city.

    Despite the fact that a substantial amount of research has been devoted to \textit{data imputation} methods, the development of models for real-time simulation of air quality stations during extended periods of hardware downtime, has not been addressed in the literature, to the best of the authors' knowledge. Therefore, this paper, proposes a SpAtial-Temporal Attention Dual LSTM (SATADL) model, which can act as an accurate proxy for stations that are offline. This paper extends the work, analyses and validation presented in \cite{Kostadinov:2025}. \colr{SATADL is an encoder-decoder deep neural network comprised of the following modules:

 \begin{itemize}
     \item \textbf{Spatial Module} - The spatial module processes the input from the surrounding stations and dynamically captures their impact on the output. It comprises of a time feature embedding block that extract time information as features and a spatial attention block that assigns weights to each surrounding station's input at every time step.
     \item \textbf{Temporal Module} - The temporal module processes the data from the offline station to extract local behaviour and patterns before the station shutoff. The module comprises of the same time feature embedding block, along with a temporal attention, that assigns weights to each time step in order to scale their impact on the output.
     \item \textbf{Decoder} - The decoder takes the outputs from the spatial and temporal modules and decodes them using two separate LSTM blocks, which are combined to produce the final output.
    
 \end{itemize}}
    
    \subsection{Related work}
    \colr{The simulation of measurement stations requires some sort of modelling of the phenomena (e.g. pollutant emission, dispersion and absorption) deemed appropriate for generating insights.}
    The topic of \colr{air quality analysis, processing and} modelling has been widely researched in the last few decades \cite{Yu:2024, Liao:2021, Baklanov:2020}. Among the effort, forecasting pollutant concentrations takes centre stage. There are two main classes of approaches in use for forecasting air quality - first-principles models and data-driven models.

    First-principle models, also referred to as physics-based or mechanistic models, such as CMAQ \cite{Eder:2006}, CHIMERE \cite{Menut:2013}, WRF-Chem \cite{Chuang:2011} and ADMS \cite{Righi:2009}, among others, model the spatial transport and deposition of different pollutants emitted from a variety of sources.
    To operate, physics-based models rely on a string of predictions, for example from regional meteorological and air pollution models, and on synthetic data, for example aggregated emission inventories \cite{Burov:2023} and low level-of-detail cadastre data. Because of their inability to directly ingest real-time measurement data, which reflects the complex local dynamics of air quality and their inherently larger operation scale, physics-based models may produce results of insufficient spatial and temporal resolution and accuracy \cite{Athira:2018}.
     
    In contrast to physics-based models, data-driven approaches seek to recreate the physics encoded in different types of observations, most commonly those of pollutant concentration and meteorological quantities. A widely used class within this family is that of statistical models which aim to learn spatial and temporal patterns and relationships in the data. %through learning patterns in relevant data.
    
    As described in \citet{Kostadinov:2025}, autoregressive integrated moving average (ARIMA) and multiple linear regression (MLR) are two very common strategies employed for data-driven forecasting. Despite their popularity, many of these methods are linear in nature, \ie\ they assume linear relationships within the time series data. In other words, they cannot be used to model complex non-linear data without extensive preprocessing or knowledge-based transformations. Additionally, these models face other limitations. For example, the standard ARIMA model is univariate and does not consider relations between variables. Variations, such as ARIMAX \cite{Jing:2009}, allow the inclusion of additional variables which have an effect on the forecasted time series. While ARIMAX models can incorporate exogenous variables, they require the values of these variables to be known in advance for accurate forecasting. In the context of air quality, future measured values are unknown and cannot be utilised.

    Multiple linear regression can handle multiple variables and a key strength is its simple training process. It assumes independent and identically distributed (i.i.d.) residuals (the differences between actual and predicted values) which is important for making valid statistical inferences about the model's parameters. \citet{Ng:2018} use meteorological factors and \pmc\ data from the previous day to forecast daily \pmc\ concentrations with MLR. In multi-step-ahead forecasting, the recursive nature of the predictions often leads to violating the i.i.d. assumption for the residuals. This is because errors in earlier predictions can propagate and accumulate in subsequent forecasts, making the residuals dependent and non-identically distributed.
    Combining the model’s inability to capture non-linear patterns with its sensitivity to the violation of statistical assumptions, limits its applicability to more complex multi-step-ahead forecasting tasks. While non-linear relationships can sometimes be addressed through domain-specific transformations or by embedding hard-coded basis functions into the model, such approaches require deep expert knowledge and may not generalize well, making them impractical or even infeasible in many real-world scenarios. As a result, standard MLR is routinely found inadequate for modelling the complex, dynamic behaviour observed in air quality time series.

    Alternative to these methods is the use of techniques from machine learning (ML) and deep learning, which offer more flexible models with potentially higher predictive power. Support vector machines (SVM), Gaussian process regression (GPR) and ensemble methods, for example, random forest (RF) are powerful ML models capable of capturing the highly complex relations between various features and thus outperforming traditional statistical methods in time-series modelling \cite{Cerqueira:2022}. Furthermore, ML models can adapt to changing conditions and exhibit greater robustness to noise and uncertainties often present in real-world data. \citet{Liu:2017} use SVM to forecast one-step-ahead daily air quality index in Beijing using meteorological conditions and air pollutants. \citet{Wang:2021} put forward a GPR model to forecast a variety of atmospheric pollutants. Under some moderate assumptions, these models provide a flexible measure of the uncertainty of the regression model itself, which is a valuable feature for sensor analysis, optimal placement and decision making. The authors explore ways to efficiently tune GPR models and find that they exhibit general applicability for air quality predictions. \citet{Ivanov:2022} use RF to predict up to seven-steps-ahead daily \pmc\ concentrations in the city of Burgas, using meteorological conditions and the last daily \pmc\ measurement. In addition, they consider the days of the year and seasons in order to find seasonal dependencies.

    As the complexity of the monitored processes increases, modelling a faulty station would require methods more flexible than those in classical ML. The members of the family of deep learning are this natural candidates for the task.
 
    Advancements in neural network (NN) architectures have allowed the development of promising models, which try to overcome many existing challenges in time-series forecasting. One type of NN, known as recurrent neural networks (RNN) has established itself as a potent solution to handling complex and long term relationships in time series. Among them, long short-term memory (LSTM), developed by \citet{Hochreiter:1997} has become the predominant choice, thanks to its solution of the vanishing gradient problem, encountered by other RNNs \cite{Bengio:1994}. Since their inception, LSTM have seen very wide usage in multiple fields, including air quality forecasting. For instance, \citet{Freeman:2018} use an LSTM to predict 8-hour-average surface ozone, \oz, concentrations and forecast up to 9 steps ahead (72 hours) in Kuwait. Their model displays significantly better results compared to ARIMA. \citet{Zhou:2019} experiment with different modifications of the LSTM with additional hidden layers and Euclidean-norm regularisation for multiple air pollutants. Unsurprisingly, the authors found that using a deeper architecture with more layers yields better results.

    Despite being sufficient on their own for short term predictions, RNN-only models are not adequate for long-term forecasting. To overcome this limitation, researchers have increasingly explored hybrid architectures that combine the strengths of different models. Other types of NN such as convolutional neural networks (CNN) are capable of extracting important insights from multivariate data. This allows for the development of compound models which can forecast many steps ahead with low error rates, while taking into consideration past trends in a multivariable environment. These hybrid models often adopt an encoder-decoder architecture, inspired by sequence-to-sequence models \cite{Sutskever:2014}. In this framework, an encoder processes the input data and transforms it into a compact, informative representation. This encoded representation is then fed to a decoder, which generates the forecasted values for multiple time steps at a time.
    Several studies have successfully implemented such approaches. For example, \citet{Du:2021} use an encoder-decoder architecture consisting of an encoder with multiple CNN layers and an LSTM decoder, to extract spatial and temporal features from air quality and meteorological data collected from multiple stations within Beijing, for multi-step-ahead forecasting of \pmf\ in the city. Similarly, \citet{Abirami:2021} proposed an encoder-decoder model for multi-step ahead forecasting of various air pollutants in Delhi, based on a CNN encoder and an LSTM decoder. They divide the city into cells, separating the stations spatially, to form a spatio-temporal approach. 
    
    Another development in NN architectures is the attention mechanism whose purpose is to give relative weights to pieces of information, thus determining which piece has the strongest impact on the result. Including attention mechanisms improves the model's ability to prioritize and focus on the most relevant features of the data, enabling it to better capture patterns and dependencies \cite{Niu:2021}. The attention weights are dynamically generated and vary based on the given input. Encoder-decoders with attention mechanisms have become powerful tools for air quality forecasting and are frequently used in the literature \cite{Feng:2023, Shi:2021, Su:2023, Zou:2021}. \citet{Huang:2021} use a spatio-attention embedded RNN to forecast hourly air quality index up to 24 steps ahead by extracting spatial and temporal dependencies of air pollutants and meteorological conditions from multiple stations in Beijing, achieving good results in long-term forecasting.

    When it comes to data imputation efforts, \citet{Wardana:2022} develop a convolutional autoencoder for a given air quality station by using available data from the surrounding stations. \citet{Kim:2021} propose a model, based on feed-forward networks only, to extract and reconstruct the trend, seasonality, bias and residuals from air quality time series. The model's main advantage is that it can do data imputations at multiple locations at once. \citet{Zhou:2017} develop a multi-layer LSTM network, which iteratively learns to fill missing values in the given input sequences. They manage to outperform multiple other benchmark models on a Beijing air quality dataset. \colr{\citet{Yu:2025a} propose a combined model that imputes time series data and forecasts at the same time, demonstrating great results in multiple real-world datasets with missing data. The novel interpolation attention, successfully imputes missing periods by considering the available spatio-temporal data, which are then given to the other components for forecasting.}
    
    However all of these methods fill gaps in existing, historical data, instead of providing a real-time proxy for a particular device. 
    \colr{It must be noted, therefore, that even though both air quality simulation and forecasting are usually performed using data-driven modelling and machine learning architectures, their objectives and operational constraints differ fundamentally. Air quality forecasting aims to predict future pollutant concentrations at a given location using only information available up to the prediction time, under strict temporal causality constraints. In contrast, data-driven air quality simulation, as considered in this work, focuses on real-time generation of pollutant concentrations using measurements from neighbouring stations and auxiliary variables. The goal of simulation is not to anticipate future states, but to reconstruct the behaviour of an offline sensor in real time, thereby maintaining spatial data continuity during temporary station shutdowns. As a result, simulation models can use information from nearby stations at the same time step, when the local data is missing - information that would not be allowed in a forecasting setting - while still operating in real time. Confusing simulation with forecasting can therefore lead to incorrect assumptions, since treating station-replacement simulation as forecasting imposes unnecessary temporal constraints and misinterprets the purpose of the model.}
    In this context, SATADL captures both historical and real-time relationships between pollutants across space and time and leverages these to not only impute missing data, but to build a virtual replacement of the defaulting station.
    
    The rest of the paper is structured as follows: \sect{M_D} presents and formulates the problem SATADL seeks to address. The architecture of SATADL, which comprises spatial and temporal modules with attention mechanisms and LSTMs and allows it to produce stable continuous results, based on previous observations and present measurements from working neighbouring sites is also described in that section. We have opted for a detailed description to enable researchers to recreate the model and all presented results. \sect{E_S} discuses the experimental setting, data cleaning, analysis and preprocessing, used during the validation of SATADL. \sect{R} analyses the acquired results and shows descriptive comparisons between the proposed model and a selection of baseline and advanced models from literature. Finally, \sect{C_F} provides the conclusion and the future direction of the research.
        
\section{Model development} \label{M_D}
\begin{figure}
        \centering
        \includegraphics[width=\linewidth]{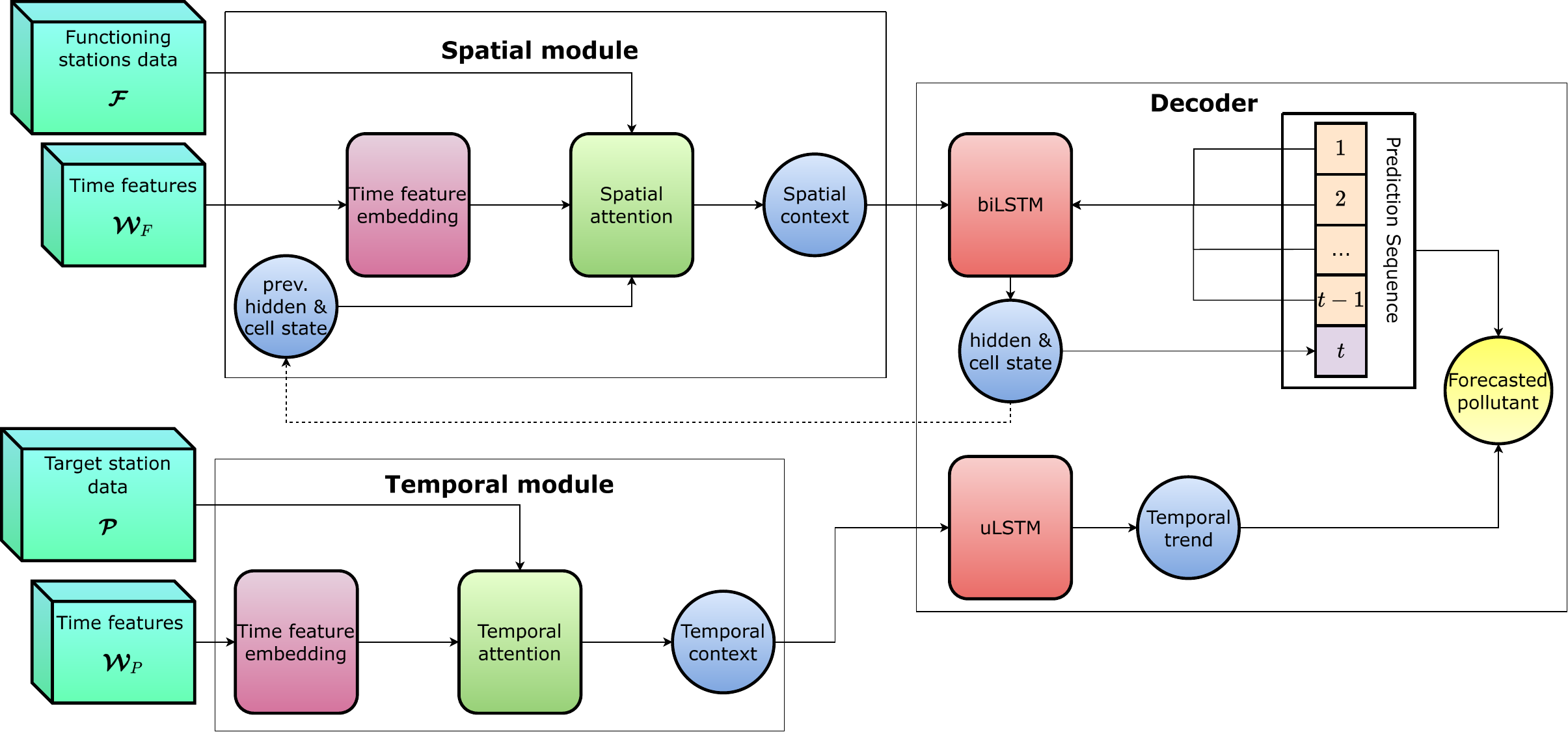}
        \caption{Model architecture diagram of SATADL.}
        \label{fig:SATADL}
    \end{figure}

In this work we expand on the proposed in \cite{Kostadinov:2025} model, by upgrading some of its components and also going in finer detail of the model's architectural design. This way we aim to provide the necessary detail for its design, while also conveying it in a replicable manner. To quickly recap - SpAtial-Temporal Attention Dual LSTM is an an encoder-decoder RNN model designed to simulate the real-time measurements of a given ``target'' station by integrating information from multiple surrounding ``functioning`` stations and historical data. The encoder consists of two modules - spatial and temporal, each with an attention mechanism. The decoder consists of two LSTM blocks, that sequentially process each of the encoder's modules outputs and combines them into a final result. An overview block diagram of SATADL is shown on \fig{fig:SATADL} and Sections~\ref{ssec:M_D_SM} to \ref{ssec:M_D_DC} are dedicated to a detailed description of each of the modules.
    
	%$
    \subsection{Problem formulation}
        We expand on the previous work, presented in \cite{Kostadinov:2025}. As a reminder to the reader, a quick setup of the problem and its formulation will be shown. Recall that we have a malfunctioned station called target that we aim to simulate with data from its historical measurements and present data from the other functioning stations. \colr{We can formally divide the time steps into two non-overlapping sets $\mathcal{T}_p$ and $\mathcal{T}_f$, which represent the past time steps up to the stopping point and the present time steps,  $|\mathcal{T}_p|=P$ and $|\mathcal{T}_f|=F$.} Our goal is to obtain a stable, continuous prediction for a chosen air pollutant at the target station, by combining the available sources of data from the different time intervals. %

        Additionally, we denote the data from the functioning stations as $\boldsymbol{\mathcal{F}}\in\mathbb{R}^{F\times N_{ff}\times N_s}$, where $N_{ff}$ and $N_s$ are the number of measured variables and number of stations, respectively. The past data from the target station is denoted as $\boldsymbol{\mathcal{P}}\in\mathbb{R}^{{P}\times N_{fp}}$, where $N_{fp}$ also represents the number of measured variables, which could differ from $N_ff$. %We would like to clarify, that although $N_{fp} = N_{ff}$ in our experiments below, this is not a limitation of our model.
        
        The last required components are the sets $\boldsymbol{\mathcal{W}}_F = \{w_i\}_{i=P+1}^{P+F}$ and $\boldsymbol{\mathcal{W}}_P = \{w_i\}_{i=1}^{P}$, where $w_i = (h_i, d_i, m_i)$ for an arbitrary index $i$, and $h_i\in\mathbb{N}_{<24}$, $d_i\in\mathbb{N}_{<7}$, $m_i\in\mathbb{N}_{<12}$ are the hour in the day, day of the week and the month, respectively, for a given data point. We refer to these as \textit{time features} for simplicity.

        With all defined variables above and a model $f$, we obtain the simulated measurements for a chosen pollutant at the target station for all $\mathcal{T}_f$ time steps.

        In \tab{tab:nom_table} we provide a brief table of notations:
\begin{table}[h]
    \centering
    \caption{Nomenclature}
    \begin{tabular}{c|p{10cm}}  % p{width} allows text wrapping
        \hline
        Name & Description \\
        \hline
        $\mathcal{T}_p$ & The set of time steps before the present moment. \\
        $\mathcal{T}_f$ & The set of time steps after the present moment. \\
        $P$ & The size of the set $\mathcal{T}_p$. \\
        $F$ & The size of the set $\mathcal{T}_f$. \\
        $\boldsymbol{\mathcal{F}}$ & The multi-dimensional array containing the data for the surrounding stations at every time step from $\mathcal{T}_f$. \\
        $\boldsymbol{\mathcal{P}}$ & The matrix containing the data for the target station at every time step from $\mathcal{T}_p$. \\
        $N_{ff}$ & The number of features each surrounding station has. \\
        $N_s$ & The number of surrounding stations. \\
        $N_{fp}$ & The number of features the target station has. \\
        $\boldsymbol{\mathcal{W}}_F$ & The matrix containing the timestamps when the surrounding measurements are taken. \\
        $\boldsymbol{\mathcal{W}}_P$ & The matrix containing the timestamps when the target measurements are taken. \\
        \hline
    \end{tabular}
    \label{tab:nom_table}
\end{table}
    
    \subsection{Spatial module}
    \label{ssec:M_D_SM}
    %$
    As described in \cite{Kostadinov:2025}, the spatial module, which takes as input $\boldsymbol{\mathcal{W}}_F$ and $\boldsymbol{\mathcal{F}}$, is designed to capture the dependencies, existing between the surrounding, functioning stations. To do this, we use an additive attention mechanism, which assigns dynamic weights, according to the impact each station has on the prediction. We refer the interested reader to \cite{Bahdanau:2016} for more details on additive attention.
    The target pollutant measurements $\{\pi_1,\pi_2, \ldots, \pi_{Ns}\}$ of each station are subsequently multiplied by the resulting weights to obtain the \textit{spatial context}, $\mathbf{sc}_t$, at time step $t$. This process is repeated $F$ times. A diagramatic representation of the spatial module is shown in \fig{fig:sm}.
    The module consists of the following key blocks:
        \begin{figure}
            \centering
            \includegraphics[width=\linewidth]{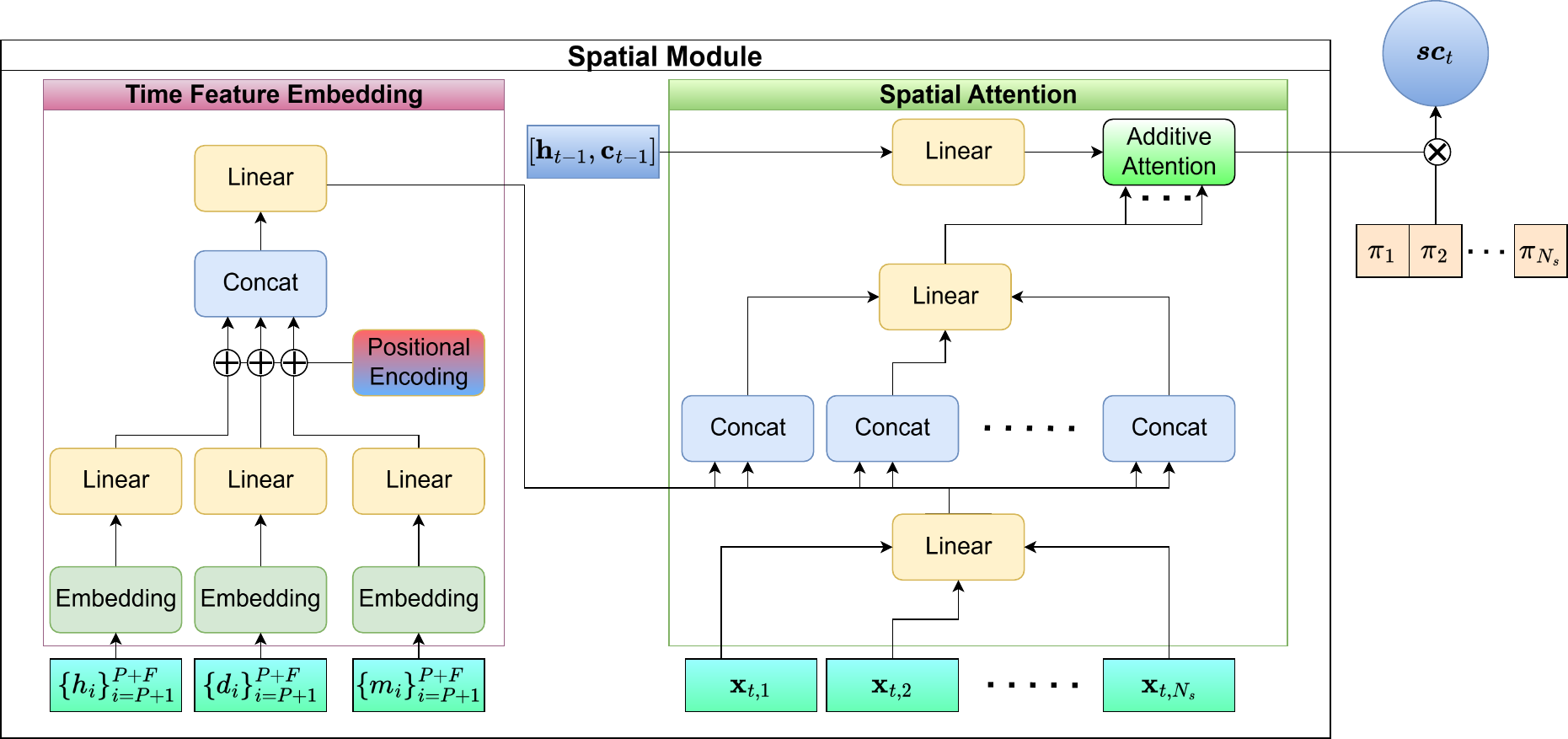}
            \caption{The spatial module of SATADL.}
            \label{fig:sm}
        \end{figure}

    \subsubsection{Time Feature Embedding}
        Embedding is a method of transforming categorical values into vectors of numbers. This is done so that the data is transformed into a format suitable for deep learning models, while also trying to preserve the connection between the unique values. For our purpose we apply embedding to the timestamps of the data to help our model learn seasonal patterns (if any are present) and how the time of the measurement affects the results. In \cite{Kostadinov:2025} the time feature embedding comprised three embedding layers, followed by a single pass through a linear layer, and concatenation. We decided to redesign the method in a way that it captures the seasonal and periodic patterns from the timestamps.%
        Each time feature, $w_i$, in the input sequence is embedded using three separate embedding layers, corresponding to $h_i, d_i$ and $m_i$, respectively.
        The embedded features are then passed through three separate linear layers. In order to achieve periodicity, we add positional encoding, denoted Pos, using sine and cosine transformations for the even and odd positions, respectively. Additionally, before applying the transformations we scale, based on the time feature's period, $\tau$
        \begin{align*}
            \text{Pos}(2i) &= \sin\left(\frac{2\pi i}{\tau}\right) \\
            \text{Pos}(2i+1) &= \cos\left(\frac{\pi(2i+1)}{\tau}\right)
        \end{align*}
        where $\tau$ changes based on the periodicity of each time feature (i.e. for the days of the week it is equal to seven).
        After all linear layers a rectified linear unit (ReLU) is applied.

    \subsubsection{Spatial attention}
        The spatial attention mechanism captures the dependency between the functioning stations and their influence on the predictions. In this way, the LSTM utilised in the decoder is enhanced by the resulting weights of the spatial attention and the weighted spatial features. Given the functioning stations' measurements at time $t$, $\mathbf{X}_t$, we use a linear layer to expand the feature vector of each station $\mathbf{x}_{t,i}$.% where $i \in \{1, 2,\ldots,N_s\}$
        The expanded feature vector is concatenated with the extracted time feature embeddings and passed through another linear layer. Additionally, the previous hidden and cell states from the LSTM block (see \fig{fig:SATADL}), responsible for the output from the spatial module ($\mathbf{h}_{t-1}, \mathbf{c}_{t-1}$) are fed back in as a tuple. To make the dimensionality of the tuple consistent with the dimensionality of the concatenated measurements, the former are passed through a linear layer. 
        Finally, the entire information is passed to the additive attention mechanism to produce weights for the influence and impact of each station for time step $t$. The target pollutant values, $\pi_{k}$, for all station stations at time step $t$ are multiplied with their respective attention weights, to form the spatial context $\boldsymbol{sc}_t$. After all linear layers, Gaussian error linear unit (GELU) activation is applied.
                
   \subsection{Temporal module}
   \label{ssec:M_D_TM}
\begin{figure}
            \centering
            \includegraphics[width=0.8\linewidth]{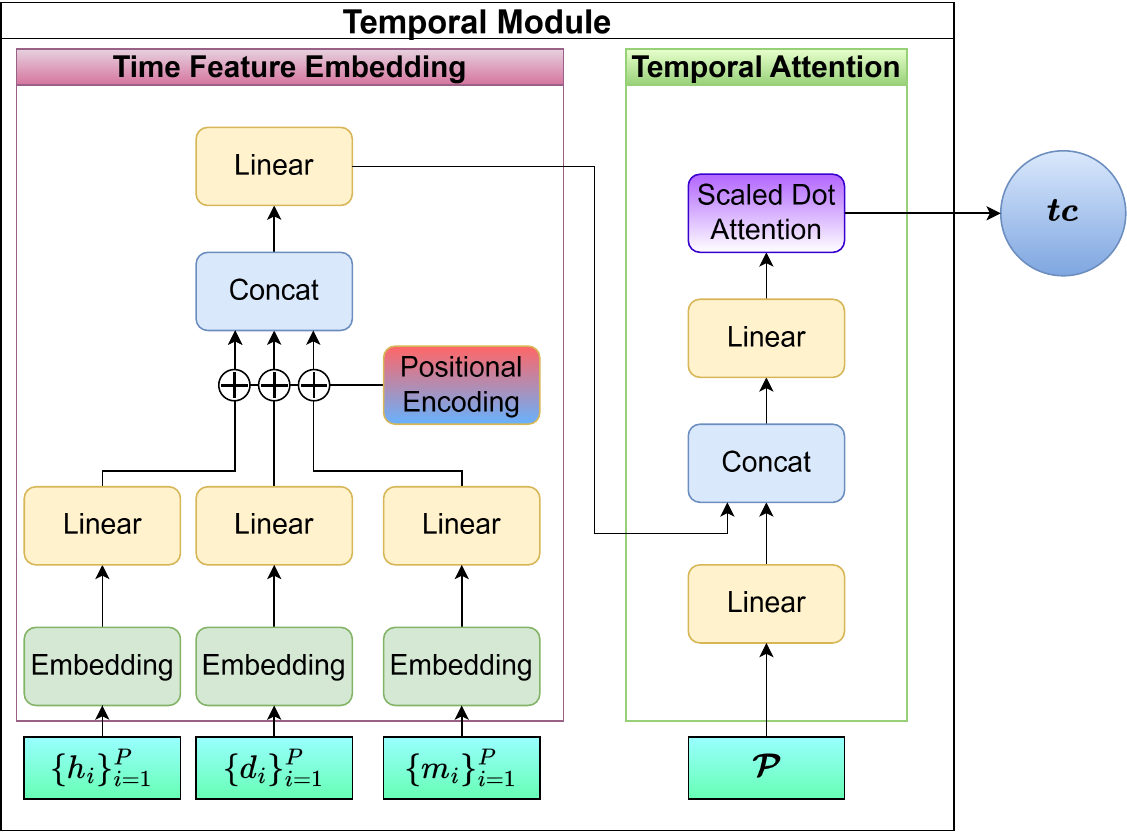}
            \caption{The temporal module of SATADL.}
            \label{fig:tm}
        \end{figure}

   %$
   The temporal module of SATADL (\fig{fig:tm}), as described in \cite{Kostadinov:2025}, captures the influence of past target-station measurements on future predictions, using $\boldsymbol{\mathcal{P}}$ and $\boldsymbol{\mathcal{W}}_P$ as inputs. Time features are embedded and combined with the measurements, followed by a scaled dot-product attention mechanism \cite{Vaswani:2017}, that forms the \textit{temporal context}, $\boldsymbol{tc}$.

    \subsubsection{Temporal attention} 
        The temporal attention operates differently than the spatial attention. Instead of finding relations and patterns between different stations, the goal of the temporal attention is to highlight significant moments in the past data from the target station and form a temporal context by assigning weighted scores to each time step. This helps the LSTM in the decoder to focus on relevant past moments, while diminishing the impact of those with little weight. Given the target station's measurements, $\boldsymbol{\mathcal{P}}$, we expand them using a linear layer. After obtaining the result, it is concatenated with the time feature embeddings and passed into another linear layer. From the output we form query, key and value matrices, with the use of 3 separate linear layers and apply scaled dot product attention to extract the temporal context $\boldsymbol{tc}$. After all linear layers we apply GELU.

    \subsection{Decoder}
    \label{ssec:M_D_DC}
\begin{figure}
            \centering
            \includegraphics[width=0.8\linewidth]{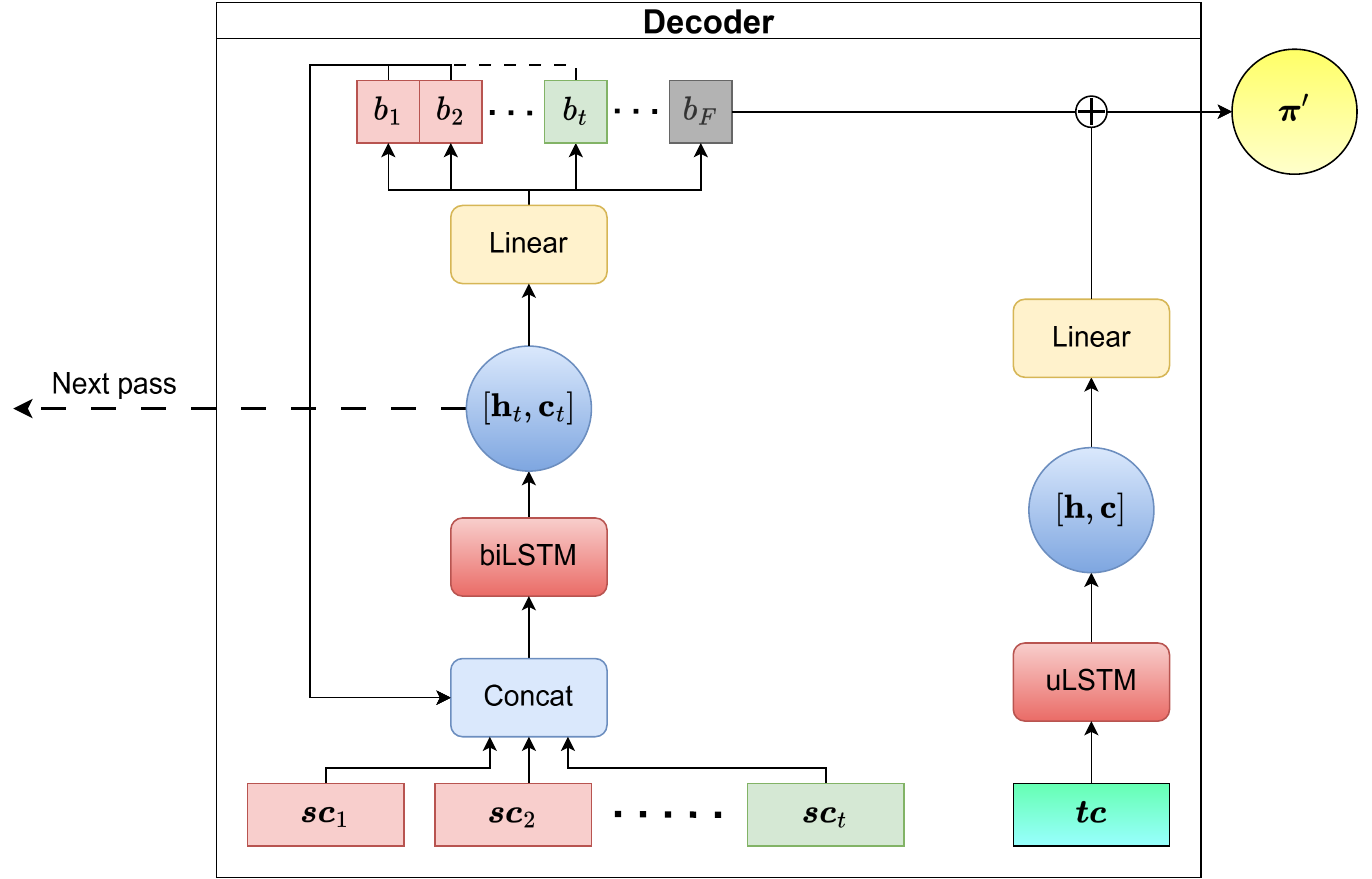}
            \caption{The decoder module of SATADL.}
            \label{fig:de}
        \end{figure}

        After obtaining the temporal context, $\boldsymbol{tc}$ and spatial contexts, up to time $t$, ${\boldsymbol{sc}_1, \boldsymbol{sc}_2, \ldots, \boldsymbol{sc}_t}$, these are passed to the decoder, shown in \fig{fig:de}.
        The decoder combines a uni-directional LSTM (uLSTM) block, which extracts local temporal patterns, and a bi-directional LSTM (biLSTM) block, which processes spatial contexts to produce base values, $\{b_1, b_2, \ldots, b_t\}$ for the target station. These base values, which represents the global pollutant level up to the current time step, only exist over $\mathcal{T}_f$ (after the target stations has ceased functioning) and so we assign them indices $1$ to $F$, for brevity. This is also reflected in \fig{fig:de}.
        When $t=F$, the base values are combined with the output of the uLSTM block to produce the final predictions for the chosen pollutant $\boldsymbol{\pi}^\prime$ (see \cite{Kostadinov:2025} for additional detail).
        
    \subsubsection{Uni-directional LSTM Block}
        After obtaining the temporal context $\boldsymbol{tc}$ from the temporal module, it is passed through an uLSTM. The final hidden and cell state ($\mathbf{h}$ and $\mathbf{c}$) from the LSTM output are then concatenated and passed through a linear layer with sigmoid activation to produce a representation of the local patterns, which serve as a trend adjustment - how the data would shift in time, based on the local patterns.

    \subsubsection{Bi-directional LSTM Block}
        This block works sequentially by adding each of the spatial contexts $\boldsymbol{sc}_{t}$ to the already extracted $\boldsymbol{sc}_1, \boldsymbol{sc}_2, \ldots, \boldsymbol{sc}_{t-1}$, increasing the sequence length each time. Then, the newly formed sequence is passed to a biLSTM layer and the final hidden and cell states ($\mathbf{h}_t$ and $\mathbf{c}_t$ are obtained. $\mathbf{h}_t$ and $\mathbf{c}_t$ are concatenated and forwarded to a linear layer to produce a single value - the base predicted value for time step $t$, $b_t$, which is appended to the previously obtained base values $b_1, b_2, \ldots, b_{t-1}$. The extracted $b_t$ is also concatenated with the final spatial context for the next time step $t+1$ allowing the decoder to directly incorporate the previous output into the next prediction step. In addition, the concatenated $\mathbf{h}_t$ and $\mathbf{c}_t$ are passed to the spatial module for subsequent time step as the new previous states. After obtaining all $b_k, \ k \in \{t_{P+1},t_{P+2},\ldots,t_{P+F}\}$, they are concatenated and summed with the output of the uLSTM block to form the final prediction for the target pollutant, $\boldsymbol{\pi}^\prime$.
        
\section{Experimental setting} \label{E_S}
In this section we describe the collection, preprocessing and analysis of four data sets of air quality measurements, used to evaluate the performance of SATADL. The data sets, which come from the cities of Beijing \cite{beijing:data}, Hong Kong\footnote{\url{https://cd.epic.epd.gov.hk/EPICDI/air/station/?lang=en}}, Sofia\footnote{\url{https://eea.government.bg/kav/}} and Delhi\footnote{\url{https://cpcb.nic.in/automatic-monitoring-data/}}, were chosen for their accessibility and relevance. We note that these datasets complement the analyses preseneted in \cite{Kostadinov:2025}, which only featured tests on air quality data from the city of Sofia.

\colr{The Hong Kong data was extracted from the Hong Kong's official Environmental Protection Department API, which was provided in \cite{Shi:2021}. The Delhi dataset was extracted from publicly available API on the official website of Central Pollution Control Board of India, provided in \cite{Abirami:2021}. Finally, the Sofia data was extracted from the publicly available API on the official website of the European Environmental Agency for Bulgaria, which has been used in \cite{Marinov:2022}. All stations were listed by the corresponding authority as being either urban or within a metropolitan area. }

\subsection{Data overview}
    An overview, including the measured variables, measurement period, number of stations and each chosen target station is shown in \tab{tab:datasets_overview}. Each pollutant is measured in \unit{\micro\gram/\meter\cubed}. The temperature (T) and dew point temperature (DPT) are measured in $^\circ\mathrm{C}$, the relative humidity (RH) in percentages and wind speed (WS) and wind direction (WD) are measured in  \unit{\meter/\s} and degrees from $\ang{0}$ to $\ang{360}$, respectively. Finally, the rainfall (R) and pressure (P) are measured in  \unit{mm} and \unit{hPa}. The choice of target station were chosen at random, because not all stations have detailed information about location, quality and other characteristics on which we can base our choice.
    %!
    \begin{table}[ht]
    \centering
    \scriptsize
    \caption{Overview of the datasets.}
    \begin{tabular}{c|c|c|c|c|c}
         \toprule
         Dataset & 
         \makecell{Number of \\ stations} & 
         \makecell{Measured \\ air pollutants} & 
         \makecell{Measured \\ meteorological \\ conditions} & 
         Period & 
         \makecell{Target \\ station} \\
         \midrule
         
         Hong Kong & 16 & 
         \makecell{\sod, \nod, \nox,\\ CO, \oz, \pmf, \pmc} & 
         - & 
         \makecell{2016-01-01 - 2019-01-10} & ``Central`` \\
         
         Delhi & 9 & 
         \makecell{NO, \nod, \nox, \\ CO, \oz, \pmf, \pmc} & 
         - & 
         \makecell{2019-01-01 - 2024-01-01} & ``Pusa`` \\ 
         
         Sofia & 5 & 
         \makecell{\sod, NO, \nod, \pmc} & 
         \makecell{T,  WS, \\ WD, RH} & 
         \makecell{2015-10-01 - 2024-01-01} & ``Nadezhda`` \\
         
         Beijing & 13 & 
         \makecell{\sod, \nod, CO, \oz, \\ \pmf, \pmc} & 
         \makecell{T, P, R, \\ DPT, WS} & 
         \makecell{2013-01-01 - 2017-03-01} & ``Huairou`` \\
         
         \bottomrule
    \end{tabular}
    \label{tab:datasets_overview}
\end{table}
    
    Initial screening of the data revealed that there are missing and negative values, as well as outliers.
    
    \tab{tab:miss_neg1} and \tab{tab:miss_neg2} show the percentage of missing and negative values of the measured pollutants and the meteorological conditions, averaged over the datasets. Attributed which are not measured in a dataset are written as `not measured` (NM). 
    
    \begin{table}[ht]
    \centering
    \scriptsize
    \caption{Missing and negative values for air quality pollutants (in \%).}
    \begin{tabular}{lcccccccc}
    \toprule
    Dataset & CO & NO & \nod & \nox & \oz & \pmc & \pmf & \sod \\
    \midrule
    Hong Kong & 45.76 / 0.01 & NM & 2.84 / 0.07 & 8.96 / 0.01 & 3.08 / 0.55 & 4.33 / 0.05 & 4.45 / 0.37 & 2.77 / 1.15 \\
    Sofia (EXEA) & NM & 2.81 / 0.29 & 2.81 / 0.24 & NM & NM & 3.85 / 0.06 & NM & 1.33 / 0.20 \\
    Delhi & 7.80 / 0.42 & 5.52 / 0.00 & 5.40 / 0.00 & 5.02 / 0.29 & 6.11 / 0.00 & 6.78 / 0.00 & 5.69 / 0.00 & NM \\
    Beijing & 4.92 / 0.00 & NM & 2.88 / 0.00 & NM & 3.15 / 0.00 & 1.53 / 0.00 & 2.08 / 0.00 & 2.14 / 0.00 \\
    \bottomrule
    \end{tabular}
    \label{tab:miss_neg1}
    \end{table}
    
    \begin{table}[ht]
    \centering
    \scriptsize
    \caption{Missing and negative values for meteorological conditions (in \%).}
    \begin{tabular}{lcccccccc}
    \toprule
    Dataset & T* & RH & WS & WD & DPT & P & R \\
    \midrule
    Sofia (EXEA) & 2.46 / -- & 3.13 / 0.00 & 8.67 / 0.00 & 8.67 / 0.00 & NM & NM & NM \\
    Beijing & 0.10 / -- & NM & 0.08 / 2.64 & NM & 0.10 / 44.26 & 0.09 / 0.00 & 0.10 / 95.98 \\
    \bottomrule
    \end{tabular}

    \vspace{0.2cm}
    \begin{minipage}{0.9\linewidth}
    \scriptsize *Negative temperature values are not reported here, as they are physically valid and not considered errors.
    \end{minipage}
    
    \label{tab:miss_neg2}
    \end{table}
            
\begin{figure}
        \centering
        \includegraphics[width=\linewidth]{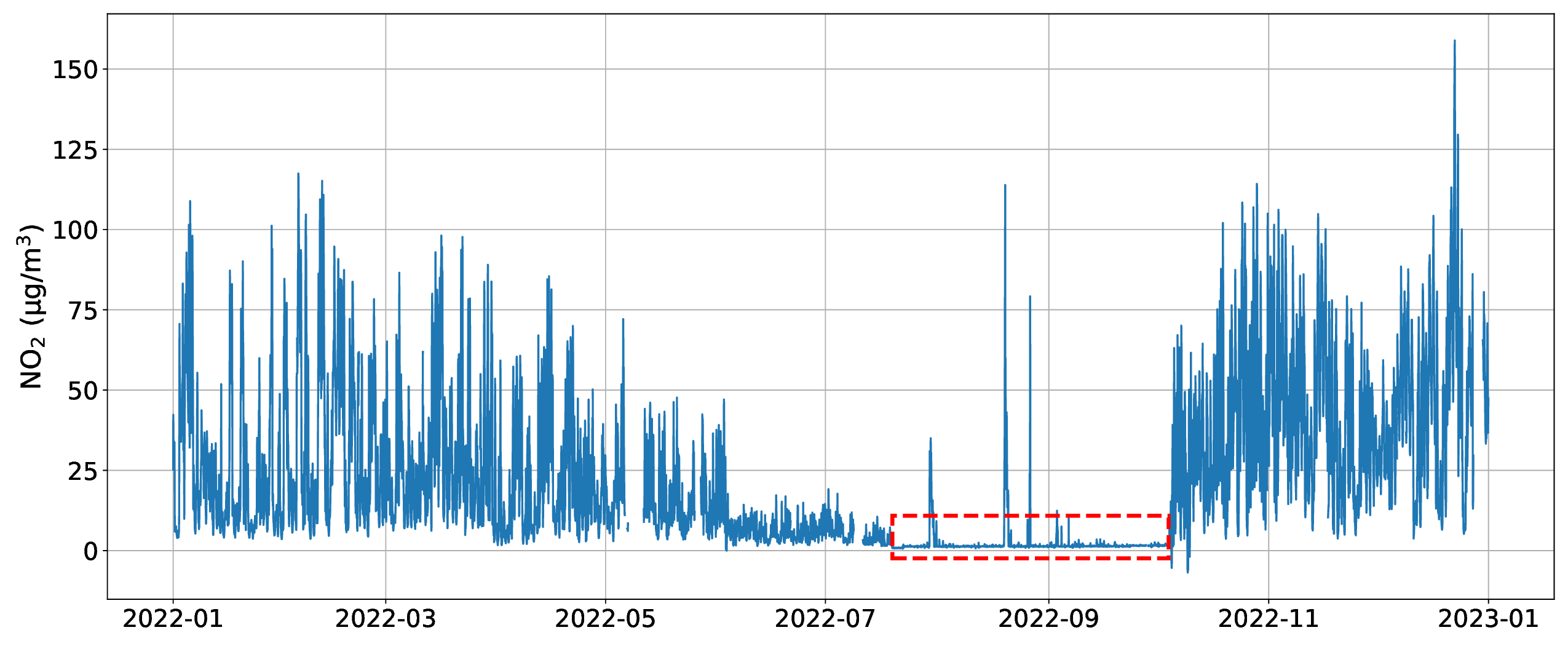}
        \caption{An example of continuous repeating values for \nod \ at Pavlovo station.}
        \label{fig:repeat_no2}
    \end{figure}

    In addition, deeper analysis in the Sofia dataset showed large periods in which measured pollutant concentrations are very similar in absolute value, but not exactly equal. These periods contain exclusively values that are much lower than those observed normally, while their difference is practically none. For instance, values of \nod, for the entire months of August through to October (red, dashed rectangle on \fig{fig:repeat_no2}), have a mean of \num{1.76} and standard deviation of \num{0.33}. We refer to the values of these periods as repeating values. The other datasets do not exhibit such anomalies.
    
    To address these problems we take the following steps:
    \begin{itemize}
        \item Negative values, except those for temperature, were treated as missing.
        \item Longer periods of repeating values were taken to indicate anomalous readings. Repeating values are marked as such based on a rolling-window standard deviation below a certain threshold. \colr{For rolling window we chose 24 hours and examined windows only when the available points were more than eight.} To the best of our knowledge, such thresholds do not exist. Based on manual inspection and distribution analysis, we determined that a threshold of 1.0 is suitable for \nod\ and \pmc, while 0.75 works well for NO. For \sod, we did not observe any large windows of repeating values. For the rest of the pollutants, not measured at the Sofia stations, we also employed 1.0 threshold but did not observe any significant repeating periods.
        Such values were present in several measured pollutants and were treated as missing data. 
        \item Large gaps in the data were removed, with experiments conducted using intervals of twenty-four, twelve, four, and one hours. Based on our findings, the best results were achieved when intervals longer than four hours were classified as large and subsequently removed. For the remaining missing data points, linear interpolation was applied.
        \item Features with the majority of data missing were excluded from consideration during model development. All other measured quantities are used as features for SATADL.
    \end{itemize}

     The final step of data cleaning involved filtering the data to retain only the timestamps where measurements are available from all stations. For instance, if one station has missing data for a certain period of time, but another has data during this period, the corresponding timestamps from the latter are also removed to ensure consistency across all stations. This ensures that the dataset contains complete records for all stations for every timestamp used. In the future, we aim to explore more into detail scenarios where multiple stations are offline. 

     By way of preprocessing the data, we apply transformation to all features, by scaling them in the range $[0,1]$ to ensure they are on a similar scale. Additionally, we extract the hour of the day, the day of the week and the month from the timestamp of the measurements and add them as separate features.

    \subsection{Stationarity analyses}
    %\subsubsection{Stationarity analyses}
    %\label{sssec:stationarity}
     When working with time series, it is important to have stationarity, since it is a prerequisite to many forecasting algorithms \cite{Marinov:2022}. Moreover, non-stationary time series are prone to distribution shifts, making them unpredictable for statistical models that rely on patterns learned from historical data. Non-stationary time series need to be stabilised during preprocessing. We conducted a stationarity check on the dataset using the augmented Dickey-Fuller test. There was no evidence suggesting statistically significant non-stationarity in any of the  attributes in all data sets.
     
\subsection{Modelling environment and setup}
    \label{ssec:mod_env}
    The training process and experimentation were conducted on a Windows 11 machine, with Intel\textsuperscript{\textregistered} Core\textsuperscript{\texttrademark} i7-1260P and NVIDIA T550 Laptop GPU. Python 3.12.6 was the main operating language including multiple packages. The data analysis, cleaning and pre-processing were done using Pandas, Numpy and Scipy. Metrics from scikit-learn were used and visualisation was concluded using Matplotlib and Seaborn.
    
    We used the proposed training setup as in \citet{Kostadinov:2025} for all datasets. While there may exist better configurations for every unique dataset, we stopped at the best found for the Sofia dataset. The proposed model was developed using Pytorch. For training, the Adam optimiser was utilised with a learning rate of 0.001 and reduction on plateau of 0.5. The loss function is set to mean squared error. The prediction steps $F$ and past steps $P$ were chosen to be 12 hours and the number of epochs was set to 30. 
    Additionally, a second round of experiments where $F$ is set to 48 hours was also conducted. In this way we want to test both short-term and long-term reliability of the simulated measurements. We chose the two values based on the missing periods and patterns, found in the datasets. The majority of periods of unavailable data are between two and eight hours, while gaps up to and larger than 48 hours were only observed in the Delhi and Sofia datasets. The number of functioning stations, $N_s$ and the number of features in each station, $N_{ff}$, are set according to \tab{tab:datasets_overview} and the subsequent data cleaning procedures. Each dataset is divided into training, validation and test sets with a ratio of 8:1:1. The full list of parameters of SATADL can be seen in \tab{tab:nn_parameters}. We used the same parameters for all datasets. We chose \pmc\ as our target pollutant, because it is measured across all station networks and carries regulatory significance. The chosen target stations can be seen in \tab{tab:datasets_overview}.
    \begin{table}[h!]
        \centering
        \caption{SATADL Parameters}
        \label{tab:nn_parameters}
        \begin{tabular}{lc}
            \toprule
            \textbf{Description} & \textbf{Value} \\
            \midrule
            Size of the time features embedding & 3 \\
            Size of the initial linear layer in the time features embedding & 9 \\
            Size of the last linear layer in the time features embedding & 1 \\
            Size of hidden state in the bidirectional LSTM & 20  \\
            Size of the linear layer, processing the spatial module input & 18  \\
            Size of the linear layer, processing the temporal module input  & 18 \\
            Size of hidden state in the unidirectional LSTM & 20 \\
            Number of layers in unidirectional LSTM & 3  \\
            Number of layers in bidirectional LSTM & 3\\
            Dropout rate in the decoder & 0.35  \\
            Dropout rate in the spatial and temporal modules & 0.35  \\
            Size of $\mathbf{Q}$, $\mathbf{K}$ and $\mathbf{V}$ in temporal attention mechanism & 18 \\
            \bottomrule
        \end{tabular}
    \end{table}
    
\section{Predictive results and discussion} \label{R}
To evaluate the performance of SATADL, in addition to the basic studies presented in \cite{Kostadinov:2025}, the model was compared to five different neural-network-based models:

\begin{itemize}
    \item \textbf{LSTM}: A single layer unidirectional long short-term memory neural network with a projection layer.
    \item \textbf{Attention-LSTM}: A model proposed by \citet{Gangopadhyay:2018}, which uses stacked LSTM layers and attention mechanism to model time series.
    \item \textbf{Transformer}: A sequence-to-sequence model developed by \citet{Vaswani:2017} that uses self-attention and cross-attention mechanisms to efficiently model complex dependencies in sequential data.
    \item \textbf{CLR}: A combined model of convolution, fully connected and recurrent layers for multivariate time series forecasting.
    \item \textbf{LSTNet}: A model proposed by \citet{Lai:2018} combining convolution and recurrent layers to extract short-term local dependency patterns among variables and to discover long-term patterns for time series trends. 
\end{itemize}
    
In terms of model implementation, we used the transformer and LSTM models available in PyTorch, the attention-LSTM and LSTNet models from the authors of \cite{Gangopadhyay:2018} and \cite{Lai:2018}, respectively and a custom code for the CLR.
To provide a fair comparison between SATADL and the selected benchmark models, we developed spatio-temporal variants of each one, comprising two copies of the respective model, one in a spatial and one in a temporal block (similar to SATADL). This step is required due to the fact that for our problem we use data from different sources and different time horizons, which need to be handled separately and then combined into a final prediction.

For clarification we refer to these models by the names of the underlying model from which they were constructed. For example, the spatio-temporal model consisting of two LSTM blocks is simply referred to as ``LSTM''.

All benchmark models were trained under the same conditions and a fixed random seed was chosen to eliminate performance variations due to non-deterministic parts of the training process. The coefficient of determination, $R^2$, and root mean squared error (RMSE) were chosen as evaluation metrics.

We conducted a twelve-hour ($F=12$) and a forty-eight-hour ($F=48$) simulation experiments with all models to test the impact of the length of period in which the target station is offline. The time it takes to train each model in the two experiments is shown in \tab{tab:training_time}.

The remainder of this section discusses the results from each experiment in detail, including the performance of different models, in light of their associated computational expense.

\begin{table}[ht]
    \centering
    \caption{Training time, in seconds, of the models considered in this study, for the 12-hour/48-hour experiments.}
    \begin{tabular}{c|c|c|c|c}
    \toprule
         Model & Sofia & Hong Kong & Delhi & Beijing \\
    \midrule
         LSTM & 283/636 & 256/593 & 116/245 & 290/717 \\
         Attention-LSTM & 304/773 & 281/594 & 128/279 & 398/924 \\
         Transformer & 1162/2044 & 1063/1981 & 579/1093 & 1023/1930 \\
         LSTNet & 502/1056 & 461/944 & 214/438 & 611/1288 \\
         CLR & 440/903 & 384/793 & 192/365 & 389/901 \\
         SATADL & 648/1589 & 443/1224 & 323/840 & 809/1737\\
    \bottomrule
    \end{tabular}
    \label{tab:training_time}
    
\end{table}

\subsection{Twelve-hour experiments}
    \label{ssec:12h}
\begin{figure}
        \centering
        \includegraphics[width=1.0\linewidth]{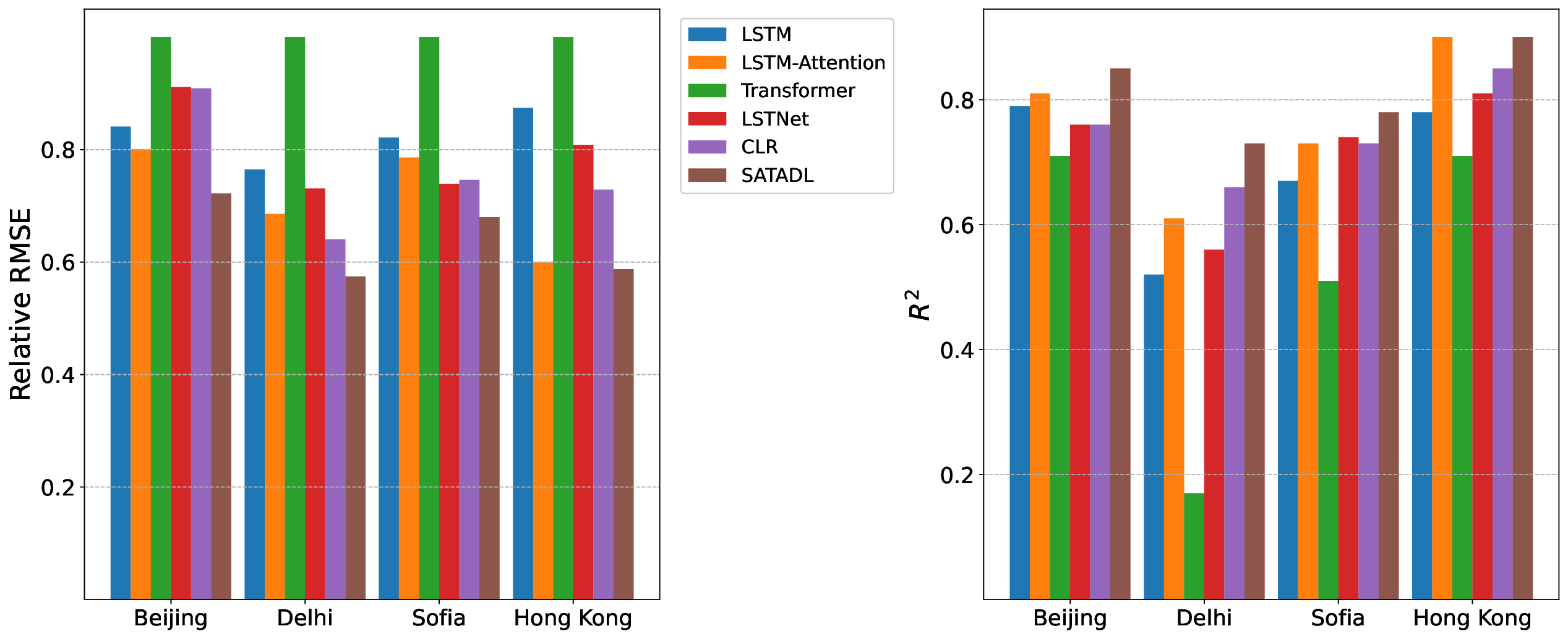}
        \caption{Root mean squared error (RMSE) and coefficient of determination ($R^2$) results for the twelve-hour simulations for each model in all datasets}
        \label{fig:12h}
    \end{figure}

    %In this section, we focus on the results, shown in \fig{fig:12h}. 
    The average performance metrics of each model, across the data sets, for the twelve-hour experiment are shown in \fig{fig:12h}. Since RMSE is dependent on the range of the \pmc\ values, which differ in the datasets, we show the scoring, relative to the RMSE of the worst-performing model.
    
    It can be seen in \fig{fig:12h} that most models are able to deal with short-term faults up to twelve hours with high $R^2$ and low RMSE. In all datasets SATADL achieves the best performance.
    
    Despite being the simplest model (see \tab{tab:training_time}), the LSTM is able to provide good results in all four datasets. The model manages to achieve the third best result in the Beijing dataset, falling behind only SATADL and attention-LSTM.
    
    The transformer exhibits the weakest performance of all models. Over the Delhi data, the model only achieves $R^2=\num{0.17}$. The transformer also took the longest to train. This might be due to the model's complex architecture and suitability for natural language tasks, rather than time series forecasting. \colr{Although variants, capable of dealing with time series data exist (for example the Informer model \cite{Zhou:2021}), we decided to focus on models with RNN architectures, to provide a direct comparison against SATADL for the simulation task. In addition, the transformer-like architectures are generally bigger (with more parameters), which makes them impractical for real-time station simulation in resource-limited scenarios. We leave an in-depth comparison with more recent transformer models for a future study.} %In the future we would like to see how more recent Transformer models for time series analysis would compete at our problem.
    
    The attention-LSTM showed strong results across all datasets. In the Beijing and Hong Kong scenarios the model achieved the second best results, while in the Delhi and Sofia it ranked as the third best. %With the addition of its very fast training time, Attention-LSTM is a suitable model for short-term simulation and there isn't a need for more complex model.
    
    The CLR and LSTNet models, which share many similarities in their architectures show mixed results. Both models contain convolutional layers that can capture local patterns. In some scenarios, for example in the Sofia data, the networks compete with SATADL for the best model, while in Beijing they fall behind the simple LSTM.
    
\begin{figure}
    \hspace*{-1cm}
    \centering
    % TikZ node wrapping the whole figure
    \begin{tikzpicture}
        % Place all subfigures inside one node
        \node[anchor=south west, inner sep=0] (fig) at (0,0) {%
            \begin{minipage}{\linewidth}
                \begin{subfigure}{\linewidth}
                    \centering
                    \includegraphics[width=\linewidth]{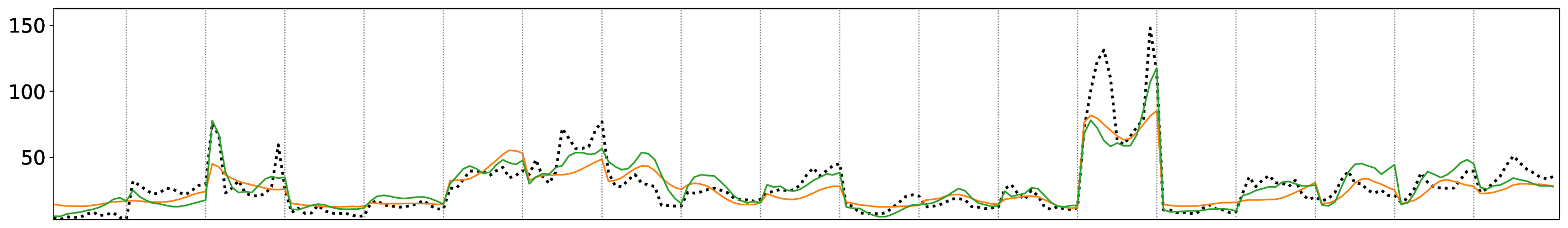}
                    \caption{}
                    \label{sfig:compare_sofia}
                \end{subfigure}
                \begin{subfigure}{\linewidth}
                    \centering
                    \includegraphics[width=\linewidth]{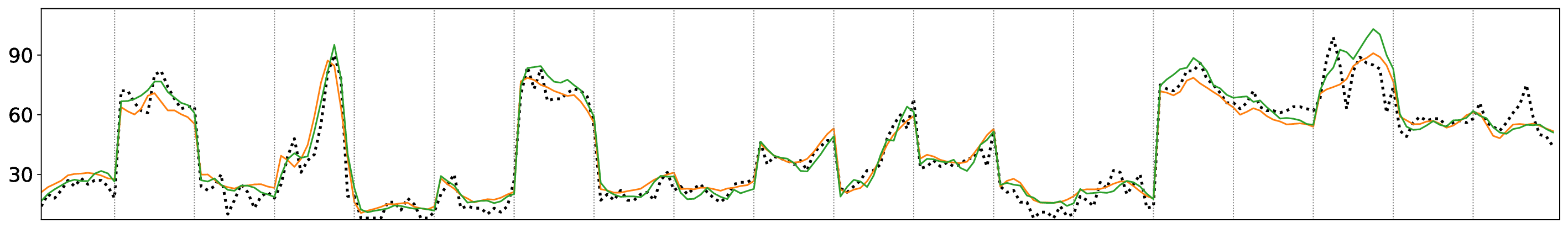}
                    \caption{}
                    \label{sfig:compare_hongkong}
                \end{subfigure}
                \begin{subfigure}{\linewidth}
                    \centering
                    \includegraphics[width=\linewidth]{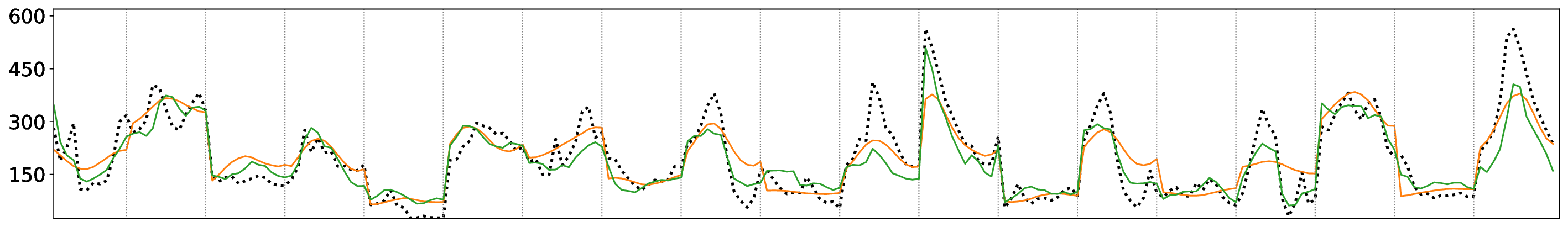}
                    \caption{}
                    \label{sfig:compare_delhi}
                \end{subfigure}
                \begin{subfigure}{\linewidth}
                    \centering
                    \includegraphics[width=\linewidth]{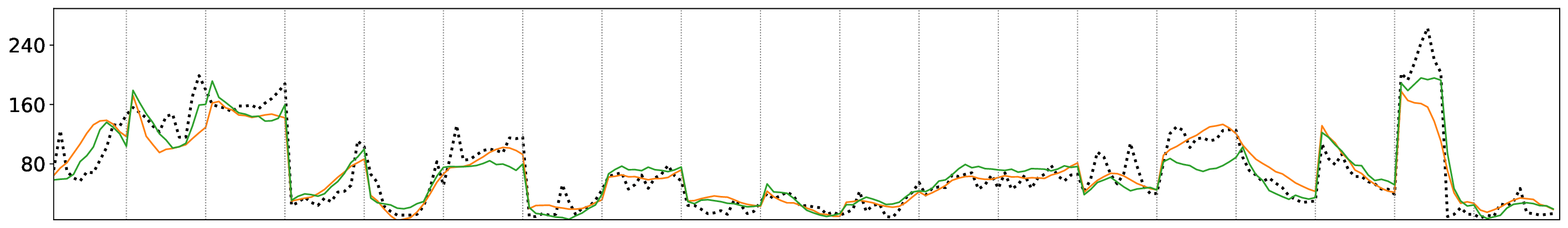}
                    \caption{}
                    \label{sfig:compare_beijing}
                \end{subfigure}
            \end{minipage}
        };

        \node[rotate=90, anchor=south] at ([xshift=-0.1cm]fig.west) 
            {$PM_{10}$ ($\mu g/m^{3}$)};
    \end{tikzpicture}

    \caption{Temporal comparison between observations (black, dotted line), SATADL predictions (green, solid line) and the second best model predictions (orange, solid line) in twelve hour simulations for (a) Sofia with LSTNet, (b) Hong Kong with attention-LSTM, (c) Delhi with CLR and (d) Beijing with attention-LSTM.}
    \label{fig:compare}
\end{figure}

    To unpack the performance statistics, we show a comparison between SATADL and the second-best model in each data set for 19 twelve-hour simulations, chosen at random (with no temporal connection) and plotted, in \fig{fig:compare}, over a continuous time axis. By choosing random periods, the models can be compared with minimal bias resulting from the choice of a particular period. In all experiments SATADL shows superior performance in capturing the qualitative nature of the observed data, compared to the runner-up models.

    Due to their simpler architecture, LSTNet (\fig{sfig:compare_sofia}) and CLR (\fig{sfig:compare_delhi}) fail to catch small variations, instead smoothing the signals from Sofia and Delhi, respectively. The models also fail to catch sharp rises or falls in the time series.
    SATADL also outperforms Attention-LSTM, over the Beijing dataset, as seen in \fig{sfig:compare_beijing}. While both models capture the general dynamics of the signal, SATADL provides closer amplitude matching, reduced bias, and better adaptation to sudden changes. Although the attention-LSTM model is prone to producing smoother predictions, it does perform similar to SATADL under favourable conditions (\fig{sfig:compare_hongkong}) and shorter simulation periods.
    Overall SATADL performs better than all benchmark models capturing sharp changes within the time series well and following the overall trend of the prediction horizon, without considerable deviation from the actual values. This is mainly attributed to the attention mechanisms of SATADL, which are able to focus on such moments in the data. The temporal module which extracts local trends and carries them through the predictions also plays an important role. Additionally, thanks to the spatial module, SATADL manages to avoid the over smoothing, evident in the predictions of the other models. In this way, the obtained results follow the observed data both locally, one hour at a time, as well as globally, over the prediction horizon.
    
    The performance gain of SATADL comes at the cost of it having the second to longest training time. Despite the unfavourable ordinal result, the difference between the fastest model (LSTM) and SATADL is just less than six minutes for substantially better results.   
    
\subsection{Forty-eight-hour experiments}
    \label{ssec:48h}
    The forty-eight-hour experiment was conducted to test the performance of SATADL for periods of longer unavailability of the target stations and to examine the stability and behaviour of error propagation as the prediction horizon, $F$, grows.
    
\begin{figure}
        \centering
        \includegraphics[width=1.0\linewidth]{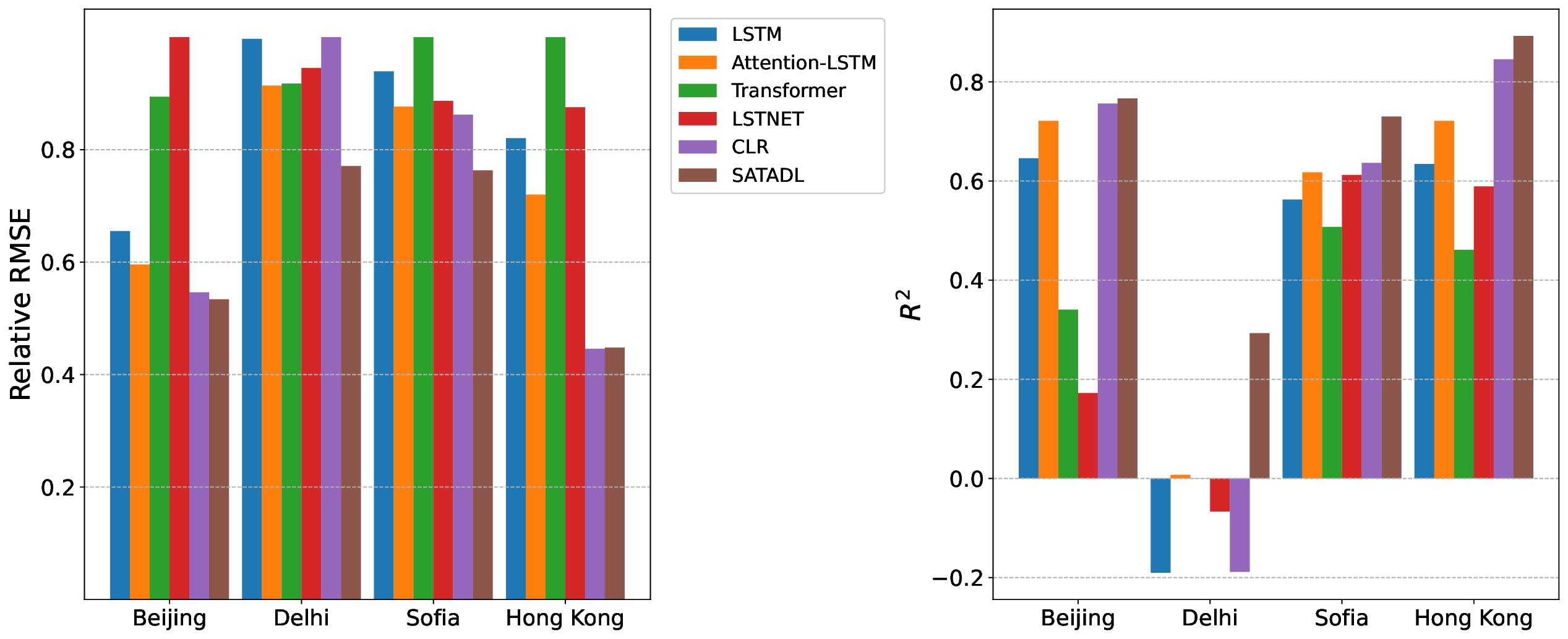}
        \caption{Root mean squared error (RMSE) and coefficient of determination ($R^2$) results for the forty-eight-hour simulations for each model in all datasets}
        \label{fig:48h}
    \end{figure}

    Similar to the results in \sect{ssec:12h}, \fig{fig:48h} shows $R^2$ and relative RMSE, averaged across predictions from hour 1 to hour 48. 
    
    \fig{fig:48h} clearly shows that none of the models performs well over the Delhi data, despite SATADL ranking highest in both metrics. 
    This result likely comes from the impact of local patterns in the data, coupled with weak correlations between the target and surrounding stations. As the time between the last measured value at the target station and the prediction point increases, the prediction drifts away faster than in the other datasets.
    
    The performance of the LSTM and the LSTNet models suffers in comparison to the twelve-hour experiments, due to their simple architecture and inability to capture and retain local and global patterns.
    
    The transformer, which ranked last in performance during the twelve-hour experiments, exhibits a moderate improvement in the Beijing and Delhi datasets, likely due to its self-attention mechanism which decreases the error growth through the prediction.
    However, the training time of the model increases to nearly 35 minutes (as compared to approximately 20 minutes for SATADL).
    
    The results of the attention-LSTM also deteriorate, showing that, notwithstanding the good performance in the twelve-hour simulations, its simple architecture is not suited for long-term simulation.
    
    The CLR again achieves mixed results. In the Hong Kong data set it ranks better on average RMSE than SATADL, while in the Delhi data the CLR performs the worst of all models. Excluding that, CLR takes a second place in terms of performance and with its relatively short training time is a good contender for a long-term simulation model.
    
    Finally, SATADL achieves substantially better average $R^2$ scores in all forty-eight-hour scenarios. The model widens its lead in the Delhi and Sofia datasets. As shown in \tab{tab:training_time}, the training time for SATADL remains high overall, but still within reasonable values to what may be required in the context of simulation under hourly measurements.
    
\subsection{Performance at specific hours}
    \label{ssec:specific_hours}
    
\begin{figure}
        \centering
        \includegraphics[width=1.0\linewidth]{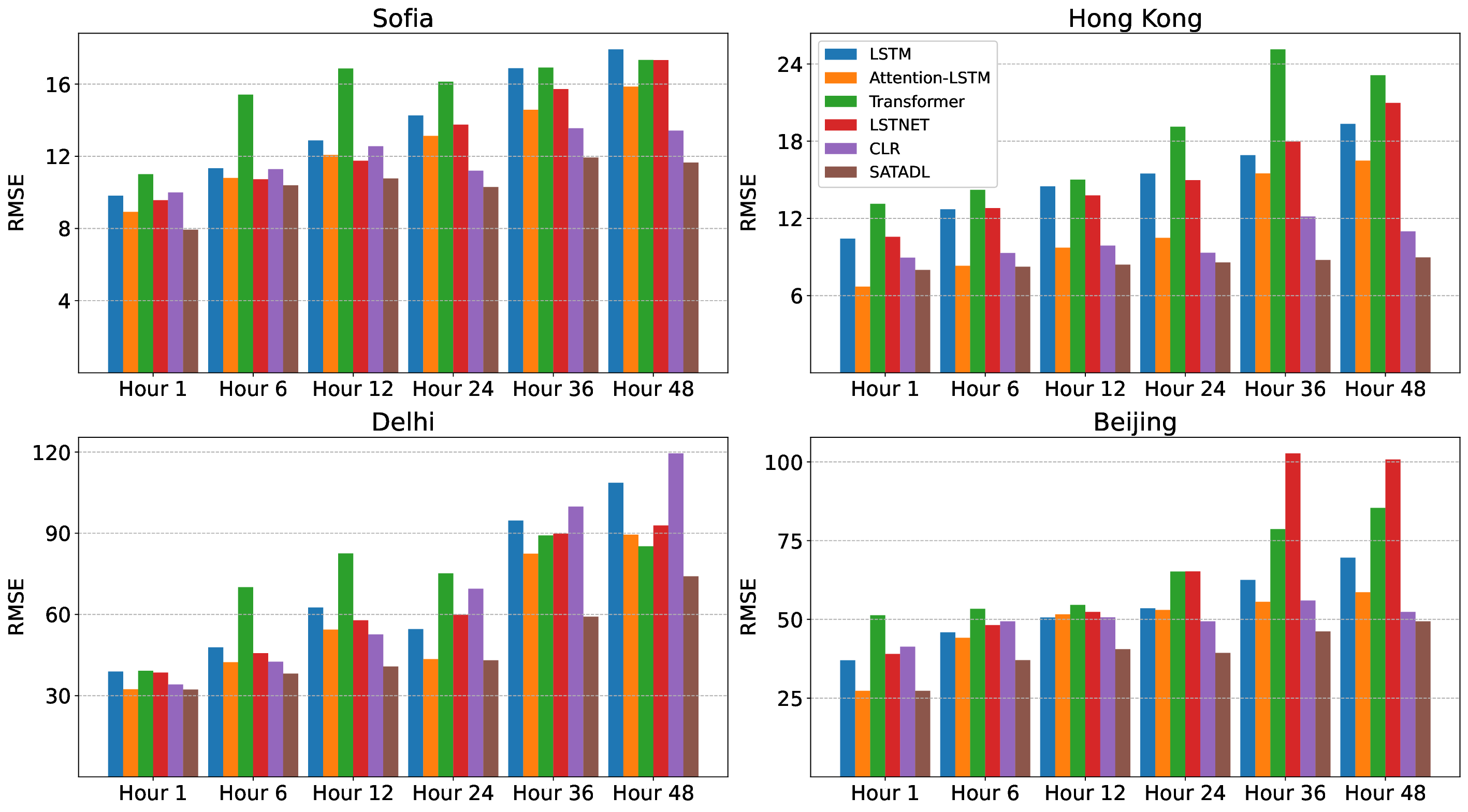}
        \caption{Detailed RMSE results for time steps 1, 6, 12, 24, 36 and 48 in the forty-eight hour simulations for every model in each dataset}
        \label{fig:rmse_hours}
    \end{figure}

\begin{figure}
        \centering
        \includegraphics[width=1.0\linewidth]{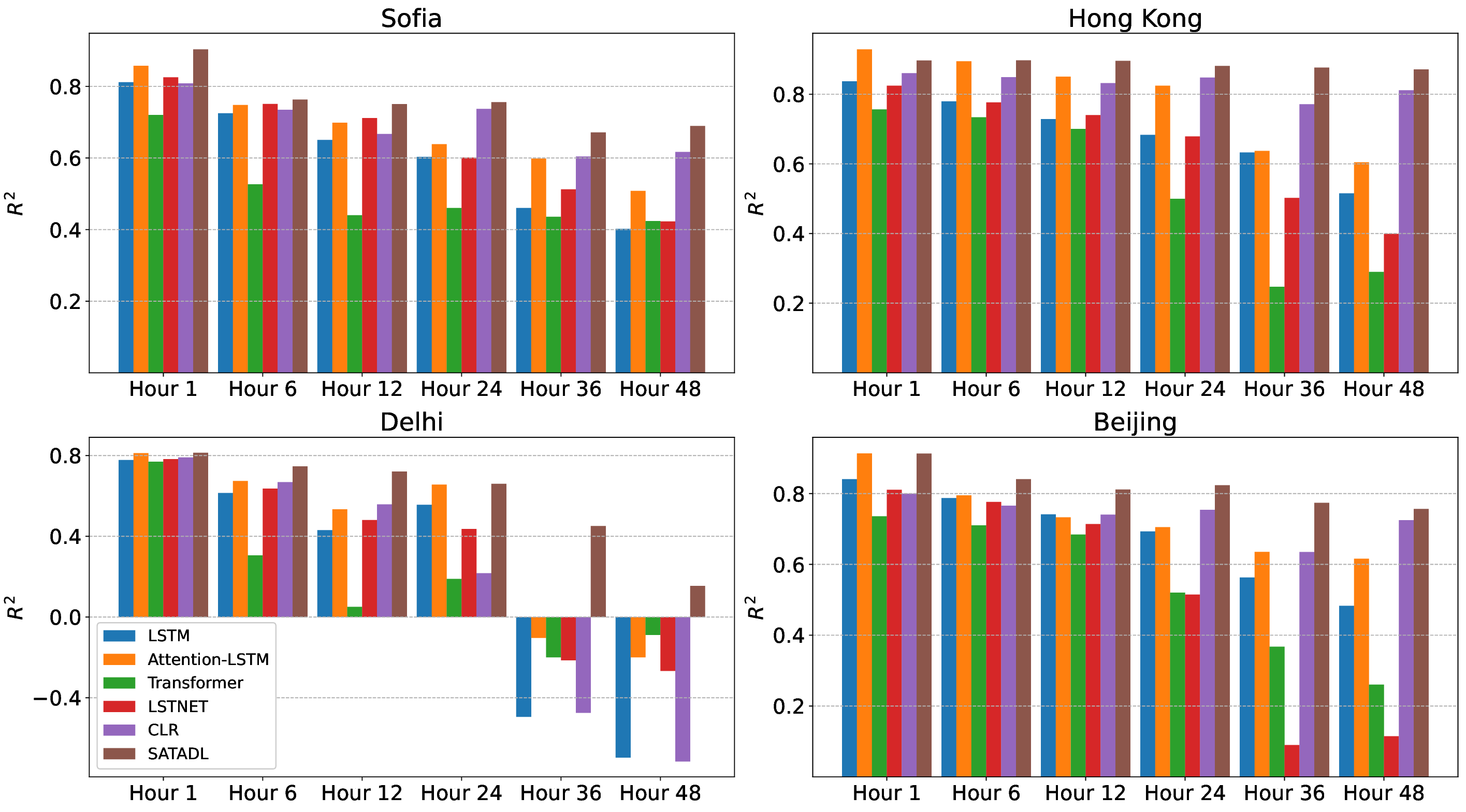}
        \caption{Detailed $R^2$ results for time steps 1, 6, 12, 24, 36 and 48 in the forty-eight hour simulations for every model in each dataset}
        \label{fig:r2_hours}
    \end{figure}

    To gain an insight on the overall performance of SATADL and the benchmark models, past that offered by averaged measures, we chose four points from the long-horizon simulations, at \num{1}, \num{6}, \num{12}, \num{24}, \num{36} and \num{48} hours and analysed $R^2$ and RMSE there. 
    The results for each data set are presented in Figures~\ref{fig:rmse_hours}~and~\ref{fig:r2_hours}, respectively.
    
    In the experiments over Sofia, Hong Kong, and Beijing, we do not see any unordinary patterns. All models gradually lose performance as the time progresses. There are some models (i.e. the transformer or LSTNet) which perform worse at hour \num{36} than \num{48}, probably due to the daily periodic patterns. The transformer performs the worst in the first time steps, but thanks to its attention mechanism, the error propagation is kept under control as the time moves forward. A similar observation can be made about the attention-LSTM, which also appears to deliver stable results. This is unlike the LSTM and LSTNet models, which lack attention in their architecture and their metrics suffer the largest decrease. Although CLR does not have an attention block, it manages to achieve consistent results overall.
    
    SATADL combines the benefits of high accuracy in the early stages with low error propagation, which results in the best performance at each time step.
    Because of the combination of information from the functioning stations at each hour and the past data and its trend, SATADL can capture the global levels of the target pollutant, while preserving the local patterns detected at the target station. As simulation time progresses, the target station local patterns are gradually lost, but thanks to the feedback mechanism in the biLSTM in the decoder, which takes as context the produced base values, $\{b_1,b_2,\ldots,b_t\}$, and the evolution of the spatial context, $\{\boldsymbol{sc}_1,\boldsymbol{sc}_2,\ldots,\boldsymbol{sc}_{t+1}\}$, it regulates the impact of the surrounding stations without drastically losing the local patterns. Additionally, the attention in the spatial module automatically removes any noisy or weakly-contributing station measurements by giving them very low weights.
    
    In the Delhi experiments we see drastic differences across all models when comparing the recent time steps with the last step. All models are able to achieve $R^2\geq0.75$, when predicting over the first hour. At later hours the models' performances significantly deteriorate. The CLR, LSTM, LSTNet and attention-LSTM record worse results than the transformer, with the first two having negative $R^2$ and sharpy increased RMSE at the later time steps.
    The only model that achieves positive $R^2$ in the last time step is SATADL, although its results are also unsatisfactory. This shows that the stations in the Delhi dataset exhibit a strong local dependency, which is not captured well by the surrounding stations.
    
    \colr{Overall, SATADL surpasses all benchmark models, delivering predictions that closely mirror real-world conditions and remain stable for up to 48 hours. This makes it a good alternative to offline physical air quality stations until they come back online. Furthermore, its compact size enables efficient deployment on edge sensor devices, with fast inference times. On average, the training takes longer than that for the other models, but it is not substantially longer. As a result, SATADL offers a reliable way to fill missing pollutant data by simulating it in real time.}

    \colr{Currently, one of the main limitations of SATADL is that it assumes that all surrounding stations operate while the target is not functioning. This may not be the case in real scenarios and we aim to experiment with multiple non-operational stations in future work. Another limitation of SATADL is that as of now, the model can only work with time series data. Static data such as building and infrastructure, environment details as well as other factors could improve the obtained results, but this would reduce the usability of SATADL in areas where such data is not available.}
    
\section{Conclusion and Future Work} \label{C_F}
In this paper we presented SATADL, a deep learning model based on attention and long short-term memory architectures, suitable for use as a virtual air quality proxy station, when the latter goes offline.

With the help of separate spatial and temporal modules, SATADL is able to extract relevant information from the functional and target stations, respectively. This information is fed into a decoder, where approximations of the real measurements are extracted with the help of two LSTM blocks. The incorporation of attention blocks that assign weights dynamically, makes the model's predictions more stable and results in larger gains between the proposed model and the benchmark ones, as the simulation horizon increases.

Using air quality data from multiple sources, different times and including different features, we performed detailed comparison between SATADL and some of the most popular deep learning models, use in air quality modelling. SATADL learns local and global patterns better, resulting in more stable and realistic behaviour of the simulated measurements and achieves better $R^2$ and RMSE results in both short- and long-term failure scenarios. This result explicitly addressed one of the future work points planned in \cite{Kostadinov:2025}, which this paper extends.

SATADL is a robust simulation model designed to address various scenarios involving data unavailability and sensor malfunctions in scenarios different to air quality modelling. For this reason, we presented the full model development procedure in the hope that others will be able to use the model in their own application-specific context. 

% What are our plans for the future?
As part of the future work, we aim to integrate the model into a live data collection system to evaluate the ability of SATADL to simulate measurements in real time when a sensor fails. This will allow us to fill in missing values and ensure uninterrupted temporal coverage across all stations. 

Additionally, we plan to enhance SATADL’s generalisation capabilities. \colr{As we stated in \sect{R}}, the model assumes that all non-target stations are fully operational; however, in real-world conditions, multiple stations may be offline simultaneously. Our goal is to extend SATADL to handle such situations effectively and to assess how prediction accuracy is affected when one or more functioning stations are unavailable. The possibility of predicting multiple quantities of interest at a time is also important to generalising SATADL. \colr{A recent paper by \citet{Yu:2025} introduces a spatio-temporal air quality prediction model which considers multi-granularity data sources. While currently not a topic of interest, we see this as a possible direction we can explore to provide more accurate predictions.}

\colr{In general, our main goals for the future of SATADL will focus on:

 \begin{itemize}
     \item Integrating SATADL into a real-time air quality data collection pipeline.
     \item Comparison against bigger transformer-based models.
     \item Development of a more adaptive solution, that deals with time periods when multiple stations are non-operational.
     \item Development of a new module that can accept static data (such as road network or building infrastructure), that influences the result.
     \item Experiment with multiple pollutant predictions.
 \end{itemize}}
Finally, we foresee the application of SATADL into scenarios more abstractly resembling physical sensor networks, for example, in increasing the reliability of heavily instrumented critical systems, which may require varying degrees of alterations to the model's architecture

\section{Declaration of generative AI and AI-assisted technologies in the writing process}
    During the preparation of this work the authors used OpenAI's ChatGPT in order to improve the readability of the text. All ideas, methods, experiments and result analysis, described in the research paper were studied and done by humans. ChatGPT was only used for writing more concise paragraphs to better convey the information. After using ChatGPT, the authors reviewed and edited the content as needed and take full responsibility for the content of the published article.
    
\section{Data statement}
    All data will be made available on request.

\end{document}